\documentclass[10pt,conference]{IEEEtran}
\IEEEoverridecommandlockouts

\usepackage{xspace}
\usepackage{tcolorbox}
\usepackage{enumitem}
\usepackage[ruled]{algorithm2e}

\usepackage{xcolor}
\definecolor{wine}{RGB}{46,84,161}

\usepackage[numbers,sort]{natbib}
\usepackage[colorlinks=true,citecolor=wine,linkcolor=wine,urlcolor=blue,bookmarks=false]{hyperref}

\usepackage{booktabs} 
\usepackage{multirow}
\usepackage{graphicx}
\usepackage{tabularx}
\newcolumntype{C}[1]{>{\centering\arraybackslash}p{#1}}
\usepackage{colortbl}
\usepackage{graphicx}  
\usepackage{float}  
\usepackage{subfig}
\usepackage{makecell} 
\usepackage{color}
\usepackage{arydshln} 
\usepackage{caption}
\usepackage{wrapfig}
\usepackage{colortbl}
\usepackage{amsthm} 
\usepackage{amsmath}
\usepackage{adjustbox}
\usepackage{threeparttable}
\usepackage{stfloats}
\usepackage{pifont}
\usepackage{xurl}
\usepackage{amsfonts}

\theoremstyle{definition}
\newtheorem{definition}{Definition}

\usepackage{microtype}
\newcommand{\ie}[0]{\textit{i.e.,}\xspace}
\newcommand{\eg}[0]{\textit{e.g.,}\xspace}

\newcommand{\tool}{\textsc{FlowScout}\xspace}

\newcommand{\todo}[1]{\textcolor{black}{#1}}

\newcommand{\hs}[1]{\textcolor{black}{#1}}

\def\BibTeX{{\rm B\kern-.05em{\sc i\kern-.025em b}\kern-.08em
    T\kern-.1667em\lower.7ex\hbox{E}\kern-.125emX}}
\begin{document}

\title{\tool: From Execution Feedback to Reliable Tool-Using Agent Workflows}

\author{\IEEEauthorblockN{
Shuo Hao\IEEEauthorrefmark{1}, 
You Lu\IEEEauthorrefmark{1},
Bihuan Chen\IEEEauthorrefmark{1},
Xin Peng\IEEEauthorrefmark{1}}
\IEEEauthorblockA{\IEEEauthorrefmark{1}College of Computer Science and Artificial Intelligence, Fudan University, China}}

\maketitle

\begin{abstract}
Agentic workflows have become an important~abstraction for building reliable LLM-based automation systems~by organizing large language models (LLMs), tools, and control~logic into explicit execution structures. However, constructing high-quality agentic workflows remains largely manual and requires substantial domain expertise. Recent studies have~explored automatic agentic workflow generation from historical~task-solving records, but they mainly produce LLM-centric workflows, where real tool executions are abstracted and simulated by LLM nodes, limiting the \hs{usability and stability} of generated workflows. To address these limitations, we propose \tool,~an execution-guided framework for generating tool-integrated agentic workflows from historical task-solving records. Specifically, \tool represents an agentic workflow as a directed graph composed of LLM nodes, tool-calling nodes, and dependency edges. It first mines \hs{a common} tool coordination skeleton from historical records to construct an initial workflow, \hs{and then refines the workflow topology through Monte Carlo tree search guided by execution feedback.}~We~evaluate \tool~on four representative task domains, and compare it with three~baselines, \ie PM4Py, ReAct and AFlow. Experimental results show that agentic workflows generated by \tool improve the tool invocation correctness by at least \hs{92.69\%} and the \hs{execution score} by at least \hs{17.66\%} over the baselines, while achieving lower~performance~variation~across~repeated~runs.

\end{abstract}

\section{Introduction}\label{sec:intro}
The rapid development of large language models (LLMs)~\cite{chatgpt, gemini} has introduced LLM agents as a new automation paradigm for complex tasks. By combining LLM reasoning with external tools, LLM agents can interpret user queries, decompose~tasks, invoke tools, and adapt subsequent actions according to intermediate results~\cite{yao2023react, shinn2023reflexion, yang2024sweagent, zhang2024autocoderover, bouzenia2025repairagent, xia2025agentless, kim2025autoresttest, schick2023toolformer, qin2024toolllm, kim2024llm}. However, the flexibility and the black-box nature of LLM reasoning make agent executions less stable and harder to inspect. For similar user queries, LLM agents may make inconsistent decisions, follow different action orders, and select different tools, leading to unpredictable execution results~\cite{anthropic2024agents}.

To improve the usability and stability of agent executions, agentic workflows provide a promising way to combine the strengths of traditional workflows and LLMs~\cite{yue2026static}. Instead~of allowing an LLM agent to freely decide every action~at~runtime, an agentic workflow organizes LLMs, tools, and control~logic into an explicit execution structure~\cite{lin2025soen101}. Such agentic workflows preserve the semantic reasoning capability of LLMs, while constraining their execution with reusable and observable structures. Therefore, agentic workflows have become an increasingly important abstraction for building reliable LLM-based automation systems~\cite{zhang2025web, he2024webvoyager, openai2025deepresearch,zheng2025mermaidflow,su2026difficulty}.

\hs{Prior studies have explored workflow generation, including synthesizing web-service workflows from formal service specifications~\cite{sirin2004htn, bertoli2010automated}, extracting process models from execution logs~\cite{friedrich2011process, van2004workflow}, and generating software workflows from natural language descriptions~\cite{mastropaolo2024github, xu2024llm4workflow}}. These studies provide important foundations for deriving workflows from specifications or~observed task executions. However, they are mainly designed for settings where task activities have explicit semantics, stable execution patterns, and well-defined interfaces. As a result, they are difficult to directly apply to open-ended tasks that require semantic understanding, dynamic decision-making, and flexible coordination among external tools. Currently, the construction of high-quality agentic workflows remains largely manual. Existing agent frameworks, \eg LangChain~\cite{langchain},~LangGraph~\cite{langgraph}, and AutoGen~\cite{wu2023autogen}, and low-code platforms, \eg Dify~\cite{dify} and Coze~\cite{coze}, provide useful abstractions for implementing agentic workflows, but developers still need to decide how the agentic workflow should be built for a target task domain, requiring substantial domain expertise and iterative debugging effort.

Recently, a few studies~\cite{zhang2025aflow, zhao2026a2flow} have started to explore~the~automatic generation of agentic workflows. These approaches employ historical task-solving records, \eg tool orchestration logs collected from human demonstrations, or LLM agent~traces where tool invocation orders have been validated by successful task completion, to generate domain-specific agentic workflows. However, they can only generate LLM-centric workflows, where nodes are all LLM operations such as planning, programming, formatting, revision, and context generation, failing to explicitly preserve the real tool invocations embedded in historical task-solving records. Thus, the generated agentic workflows still rely heavily on LLMs to simulate tool execution, which may lead to unstable behaviors, hallucinated outputs, and limited interpretability of domain-specific tool coordination.

To address these limitations, we propose \tool,~an execution-guided framework for automatically generating tool-integrated agentic workflows from historical task-solving records. Specifically, we model an agentic workflow as a directed graph composed of LLM nodes, tool-calling nodes, and dependency edges that encode data flow and control flow, and formulate its generation as a search problem over candidate workflow graphs. \hs{Given available tools and a set of historical task-solving records consisting of domain-specific user queries, reference execution results, and validated tool orchestration logs, \tool first mines the most common tool coordination skeleton from historical records, and constructs an initial agentic workflow by augmenting the skeleton with necessary LLM nodes. Then, \tool performs execution-guided graph search, where candidate workflows are executed on historical user queries and their outputs are compared against reference results. The feedback guides Monte Carlo tree search~\cite{browne2012survey} to refine the workflow graph by adding, removing or refining LLM and tool-calling nodes, as well as modifying dependency edges. Finally, \tool produces a reusable and stable agentic workflow that explicitly coordinates real tools for solving new user queries in the target~domain.}

We have conducted extensive experiments to evaluate the effectiveness and efficiency of \tool. First, we evaluate \tool on four representative task domains in ToolBench~\cite{qin2024toolllm}, \ie finance, sports, travel, and weather, and compare it with three baselines, \ie PM4Py~\cite{van2004workflow}, ReAct~\cite{yao2023react}, and AFlow~\cite{zhang2025aflow}. The results show that agentic workflows generated by \tool improve the tool invocation correctness by at least \hs{92.69\%} and the \hs{execution score} by at least \hs{17.66\%} over~the baselines, \hs{while reducing the coefficient of variation across repeated runs by \hs{62.09\%} and \hs{27.90\%} compared with ReAct and AFlow, respectively.} Second, \hs{the workflows generated by \tool incur at least 24.12\% higher runtime cost than ReAct and AFlow for improved effectiveness. Besides, ablation studies show the contribution of our workflow miner and Monte Carlo tree search in \tool. Finally, we demonstrate the generalization capability of agentic workflows generated by \tool by deploying them at runtime with tools and LLMs different from those used during workflow~generation.}

The main contributions of this work are as follows:
\begin{itemize}[leftmargin=*]
    \item We represent an agentic workflow as a structured graph consisting of LLM nodes, tool-calling nodes, \hs{and edges that encode data dependency or control dependency}, and cast the generation of agentic workflows as a graph search problem. 
    \item We propose \tool, an execution-guided generation framework, that automatically searches for tool-integrated agentic workflows based on historical task-solving records, improving the usability and stability of generated workflows.
    \item We implement a prototype of \tool, and conduct experiments to demonstrate its effectiveness and efficiency.
\end{itemize}

\section{Problem Formulation}
\hs{Given available tools and historical task-solving records, including user queries, reference execution results, and validated tool orchestration logs, we aim to automatically generate tool-integrated agentic workflows for autonomous task completion. We first define such workflows and then formulate~their~generation as a search problem over valid workflow graphs.}

\subsection{Workflow Definition}\label{sec:definition}

Given a task domain with a set of available tools $\mathcal{T}$, we define a tool-integrated agentic workflow as a directed graph.

\begin{definition}[Tool-Integrated Agentic Workflow] A tool-integrated agentic workflow is a directed graph $\mathcal{G} = (\mathcal{V}, \mathcal{E})$, where $\mathcal{V}$ is a set of workflow nodes consisting of LLM nodes and tool-calling nodes, and $\mathcal{E}$ is a set of directed dependency edges that define the control flow and data flow among nodes.
\end{definition}

\begin{definition}[LLM Node] An LLM node $v_l \in \mathcal{V}$, which performs semantic reasoning on textual context, is~a~tuple $v_l = \langle \delta_{v_l}, \rho_{v_l} \rangle$, where $\delta_{v_l}$ is an LLM and $\rho_{v_l}$ is its prompt.
\end{definition}

\begin{definition}[Tool-Calling Node] A tool-calling node $v_t \in \mathcal{V}$, which invokes real tools, is a tuple~$\langle t, \vartheta_{t} \rangle$,~where $t \in \mathcal{T}$ is the tool and $\vartheta_{t}$ is the arguments for tool invocation.
    
\end{definition}

\begin{definition}[Dependency Edge] An edge $e_{\langle v^i, v^j \rangle} \in \mathcal{E}$ is a tuple $\langle v^i, v^j, \mu \rangle$, denoting a directed dependency from node $v^i$ to node $v^j$, where $\mu$ specifies the dependency semantics, \ie data dependency or control dependency.
\end{definition}

Workflow execution starts from an LLM node for user query parsing, and proceeds along dependency edges in the graph until a final response is produced by another LLM node for \hs{response synthesis.} During execution, LLM nodes reason over available context and intermediate results, whereas tool-calling nodes provide external capabilities through real tool invocations.

\subsection{Workflow Generation as Graph Search}\label{sec:formulation}

The input consists of a set of available tools $\mathcal{T}$ and a collection~of~historical task-solving records $\mathcal{D}=\{(q_i, r_i, \pi_i)\}_{i=1}^{N}$, where $N$ is the number of records, $q_i$ is a user query, $r_i$ is the reference execution result, and $\pi_i$ is the validated tool orchestration log. These records support both the automatic construction and evaluation of tool-integrated agentic workflows.

Based on the definition in Sec.~\ref{sec:definition}, we define the search space of candidate workflows over available tools $\mathcal{T}$ \hs{by Eq.~\ref{eq:search_space}},
\begin{equation}
\label{eq:search_space}
\small
\Omega(\mathcal{T}) = \left\{ \mathcal{G}=(\mathcal{V},\mathcal{E})
\mid \mathrm{Valid}(\mathcal{G},\mathcal{T})=\text{True} \right\}
\end{equation}
where $\mathrm{Valid}(\mathcal{G},\mathcal{T})$ is a validity predicate indicating whether~\hs{a candidate workflow $\mathcal{G}$ satisfies the basic constraints of a tool-integrated workflow. Specifically, $\mathcal{G}$ is valid only if its tool-calling nodes can invoke real tools in $\mathcal{T}$, its dependency edges form an executable graph, and it contains at least one complete execution path from the input query to the final response.}

To estimate the quality of a candidate workflow $\mathcal{G}\in\Omega(\mathcal{T})$, we execute $\mathcal{G}$ on each historical query $q_i$ in $\mathcal{D}$, producing a tool orchestration trace $\hat{\pi}_i$ and an execution result $\hat{r}_i$. \todo{Since complex tool-use tasks require both correct tool selection and execution scheduling~\cite{xu2026tps}, we evaluate candidate workflows at both the tool-invocation level and the final-result level.} Specifically, we compare $\hat{\pi}_i$ with the validated orchestration log $\pi_i$ to evaluate \textit{tool invocation correctness}, and compare $\hat{r}_i$ with the reference execution result $r_i$ to obtain \textit{execution score}.

\textit{Tool Invocation Correctness (TC)} measures whether the~agentic workflow invokes the correct tools with correct~arguments while avoiding redundant invocations. Given~a~historical query $q_i$, let its validated tool orchestration log be $\pi_i=\{(t_i^k,\theta_i^k)\}_{k=1}^{K_i}$, \hs{where $K_i$ is the number of tool~invocations, and $t_i^k$ and $\theta_i^k$ denote the tool name and arguments of the $k$-th reference invocation.} Similarly, let the tool orchestration trace produced by a candidate workflow $\mathcal{G}$ be $\hat{\pi}_i=\{(\hat{t}_i^k,\hat{\theta}_i^k)\}_{k=1}^{\hat{K}_i}$, where $\hat{K}_i$ is the number of tool invocations, and $\hat{t}_i^k$ and $\hat{\theta}_i^k$ denote the tool name and arguments of the $k$-th invocation. We define the tool-invocation correctness of $\mathcal{G}$ on query $q_i$ by~Eq.~\ref{eq:tc_query},
\begin{equation}
\small
\label{eq:tc_query}
TC_i(\mathcal{G})=
\hs{\frac{
\left|
\operatorname{LCS}_{t=\hat{t}\land \theta=\hat{\theta}}
(\pi_i,\hat{\pi}_i)
\right|
}{
\max(K_i,\hat{K}_i)
}}
\end{equation}
\hs{where $\operatorname{LCS}_{t=\hat{t}\land \theta=\hat{\theta}}(\pi_i,\hat{\pi}_i)$ denotes the longest common subsequence between the reference tool invocations and the workflow-produced tool invocations, where matched invocations must have identical tool names and arguments. The numerator counts the number of correctly matched tool invocations.} Compared with normalizing by $K_i$, the denominator $\max(K_i,\hat{K}_i)$ penalizes both missing and redundant tool invocations. The overall tool invocation correctness of $\mathcal{G}$ on historical records $\mathcal{D}$ is then defined by Eq.~\ref{eq:tc_overall},
\begin{equation}
    \small
    \label{eq:tc_overall}
    TC(\mathcal{G}) = \frac{1}{N} \sum_{i=1}^{N} \mathrm{TC}_i(\mathcal{G})
\end{equation}
where $N$ is the number of historical user queries in $\mathcal{D}$.

\textit{Execution Score (ES)} measures whether the final execution result produced by the agentic workflow satisfies the user query and matches the reference execution result. \hs{Since the execution result may be expressed as natural language, structured results, or task-completion outcomes,} rule-based evaluation is often insufficient. Therefore, we adopt an LLM-as-a-judge evaluator, \ie G-eval~\cite{liu2023g}, to assess the final execution result.

Given a historical query $q_i$ with reference execution result~$r_i$ and \hs{the final execution result} $\hat{r}_i$ produced by~candidate workflow $\mathcal{G}$, the evaluator scores $\hat{r}_i$ against $r_i$ from dimensions
$\mathcal{X}=\{\textit{correctness}, \textit{completeness}, \textit{relevance}, \textit{clarity}\}$. Let $s_i^x\in[0,1]$ denote the score of $\hat{r}_i$ on dimension $x\in\mathcal{X}$. We define the execution score of $\mathcal{G}$ on historical records $\mathcal{D}$ by~Eq.~\ref{eq:as},
\begin{equation}
    \small
    \label{eq:as}
    ES(\mathcal{G})=\frac{1}{N} \sum_{i=1}^{N} \frac{1}{|\mathcal{X}|} \sum_{x\in\mathcal{X}} s_i^x
\end{equation}
where $N$ is the number of historical user queries in $\mathcal{D}$.

The objective of tool-integrated agentic workflow generation is to identify a valid workflow graph that achieves high \hs{tool invocation} correctness and execution score on historical task-solving records. Given the search space $\Omega(\mathcal{T})$, we define~the optimal tool-integrated agentic workflow $\mathcal{G}^{*}$ on tools $\mathcal{T}$~by~Eq.~\ref{eq:objective}.
\begin{equation}
\label{eq:objective}
\mathcal{G}^{*}
=
\arg\max_{\mathcal{G}\in\Omega(\mathcal{T})}
\left(TC(\mathcal{G}) + ES(\mathcal{G})\right)
\end{equation}
This formulation casts workflow generation as a graph search problem over the constrained workflow space $\Omega(\mathcal{T})$.

\section{Methodology}\label{sec:approach}

\begin{figure}
    \centering
    \includegraphics[width=\linewidth]{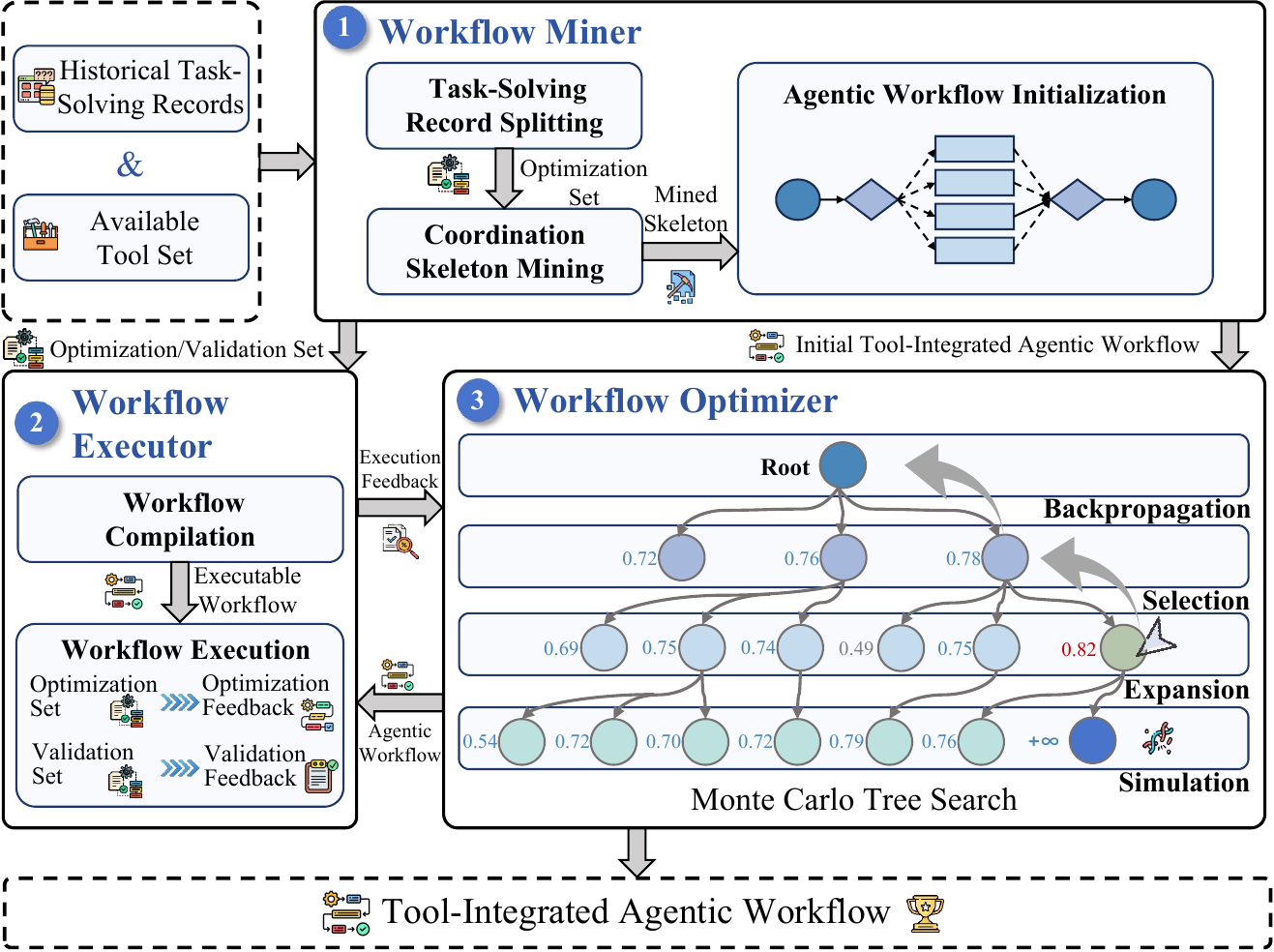}
    \caption{Approach Overview of \tool}
    \label{fig:overview}
\end{figure}

As formulated in Sec.~\ref{sec:formulation}, tool-integrated agentic workflow generation can be viewed as a graph search problem over the constrained workflow space. However, directly searching this space is inefficient because many graph edits may lead to invalid or ineffective workflows. Therefore, \tool combines skeleton mining with execution-guided graph search to generate stable and reusable tool-integrated agentic workflows.

Fig.~\ref{fig:overview} shows the overview of \tool, which consists of three modules, \ie \textit{Workflow Miner}, \textit{Workflow Executor}, and \textit{Workflow Optimizer}. Given historical task-solving records and available tools, \tool first splits the records~into optimization and validation sets, mines the frequent tool coordination skeleton from validated orchestration logs \hs{in the optimization set}, and constructs an initial tool-integrated candidate workflow by augmenting the mined skeleton with necessary LLM nodes (see Sec.~\ref{sec:workflow_miner}). Then, \tool compiles the candidate workflow into an executable form, and runs it on the optimization set and validation set to collect \hs{optimization feedback and validation feedback} (see Sec.~\ref{sec:workflow_executor}). Based on this \hs{execution feedback}, \tool performs Monte Carlo tree search to refine the topology of the agentic workflow through selection, expansion, simulation, and backpropagation (see Sec.~\ref{sec:workflow_optimizer}). After iterative optimization, \tool outputs the candidate workflow with the best validation results as the final tool-integrated agentic workflow.

\subsection{Workflow Miner} \label{sec:workflow_miner}
Directly searching for agentic workflows from an empty graph is inefficient, because most random graph edits either produce invalid workflows or fail to capture the common tool invocation structure in the target domain. Therefore, \tool first splits historical task-solving records, mines the tool coordination skeleton, and initializes a workflow $\mathcal{G}^0$, providing a structured warm start for subsequent optimization.

\textbf{Task-Solving Record Splitting.}
Given the historical task-solving records $\mathcal{D}=\{(q_i, r_i, \pi_i)\}_{i=1}^{N}$, \tool splits them into an optimization set $\mathcal{D}_{opt}$ and a validation set $\mathcal{D}_{val}$. $\mathcal{D}_{opt}$ is used to mine the tool coordination skeleton and provide optimization feedback for workflow generation, while $\mathcal{D}_{val}$ is reserved for selecting the best candidate workflow. 

\textbf{Coordination Skeleton Mining.}
After obtaining the optimization set $\mathcal{D}_{opt}$, we mine a process model from validated orchestration logs to obtain a reusable tool coordination skeleton that can serve as the template for initializing the agentic workflow. For each validated orchestration log $\pi_i={(t_i^k,\theta_i^k)}_{k=1}^{K_i}$ from historical task-solving records \hs{in $\mathcal{D}_{opt}$}, we derive the ordered tool sequence $\tau_i=(t_i^1,t_i^2,\ldots,t_i^{K_i})$, which preserves tool dependencies such as execution order and~repeated~invocations.

Based on the ordered tool sequences, we construct an event log for process mining. \hs{In the event log, $\tau_i$ is treated as a case, and each tool invocation $t_i^k$ in $\tau_i$ is treated as one activity in the corresponding case.} Then, we apply PM4Py~\cite{van2004workflow} to discover the most common process model from the constructed event log. The \hs{mined} model summarizes how~tools are commonly coordinated in most historical executions, \ie~sequential relations, alternative paths, parallel relations, and repeated~invocations. 

Then, we convert the discovered process model into a reusable tool coordination skeleton. Each activity in the mined model is mapped to a generic tool-calling position, rather than being bound to a concrete tool. This abstraction indicates that a tool invocation should be performed at this position, while leaving the concrete tool selection and argument generation to the later workflow execution. Each process relation is mapped to a workflow dependency. Specifically, sequential relations are converted into ordered dependencies between tool-calling positions, alternative paths are converted into conditional branches, parallel relations are converted~into~parallel workflow branches, and repeated invocations are converted into retry~loops. 

\begin{figure}[t]
    \centering
    \includegraphics[width=\linewidth]{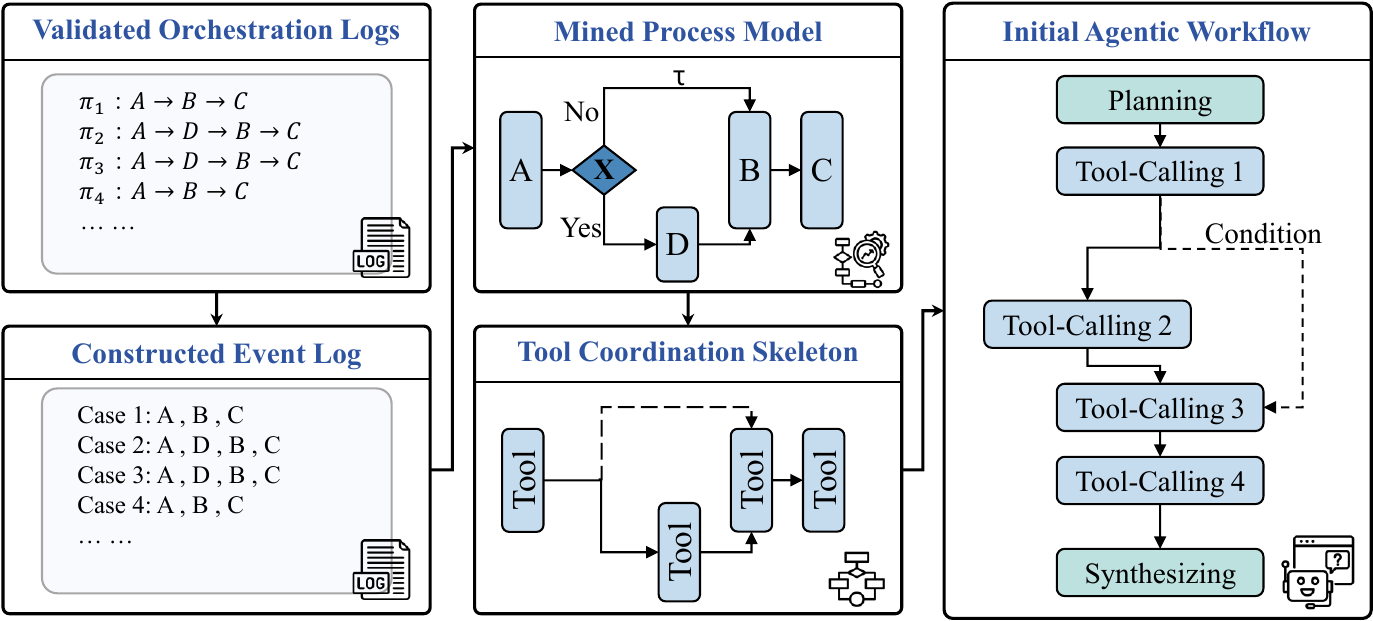}
    \caption{Example of Initial Agentic Workflow Mining}
    \label{fig:initial_workflow_mining}
\end{figure}

\textbf{Agentic Workflow Initialization.}
The resulting coordination skeleton captures the common tool-invocation structure in historical executions, but it only describes the abstract positions of tool invocations and their coordination relations. To make the skeleton executable for new user queries, we further instantiate it as an initial tool-integrated agentic workflow. Specifically, each generic tool-calling position in the skeleton is instantiated as a tool-calling node, which will select a concrete tool from the available tool set and generate arguments during execution. We further insert an LLM node for planning before the tool-calling structure to parse the user query, and an LLM node after the tool-calling structure for \hs{synthesizing} the final response to the user. Fig.~\ref{fig:initial_workflow_mining} shows an example of the workflow mining~process.

\subsection{Workflow Executor}\label{sec:workflow_executor}

The workflow executor turns a candidate tool-integrated agentic workflow into an executable program, and collects execution results. Since candidate workflows produced by miner or optimizer may be incomplete or ineffective, \tool executes them under an observable runtime, and uses the observed \hs{feedback} to guide subsequent workflow optimization.

\textbf{Workflow Compilation.}
Given a candidate workflow $\mathcal{G}'= (\mathcal{V}', \mathcal{E}')$, \tool compiles it into an executable Python script $\hat{\mathcal{G}'}$. Specifically, \tool traverses the workflow graph, and generates the corresponding Python code for nodes $\mathcal{V}'$ and edges~$\mathcal{E}'$. Each LLM node $v_l \in \mathcal{V}'$ is assigned~a function that calls the backbone LLM $\delta_{v_l}$ with certain prompts $\rho_{v_l}$~to obtain the response. \hs{Each tool-calling node $v_t \in \mathcal{V}'$ is compiled into a function that performs tool selection and argument generation when the node is executed. Specifically, the function selects a concrete tool $t \in \mathcal{T}$ based on the intermediate context, and fills the arguments $\vartheta_t$ to invoke the tool, recording the returned observation.} Dependency edges in $\mathcal{E}'$ are compiled into the data and control flow among these functions.

During compilation, we instrument each $v_l$ and $v_t$ with observation hooks. \hs{For a query $q_i$, the runtime records the input, output, and possible error message of each executed node as node-level observations $\mathcal{O}_i$. For an LLM node, the input consists of the intermediate context generated by preceding nodes and the node's system prompt, while the output is the model response passed to downstream nodes or returned to the user.
For a tool-calling node, the input consists of the tool-oriented subtask decomposed by preceding nodes and the set of available tools $\mathcal{T}$. Its output includes the selected tool, the generated arguments, and the tool response passed to downstream nodes.} These observations make candidate execution traceable, allowing \tool to identify failures such as wrong tool selection, incorrect argument generation, skipped branches, failed tool calls, or poor response synthesis.

\textbf{Workflow Execution.}
After compilation, we execute $\hat{\mathcal{G}'}$~on both $\mathcal{D}_{opt}$ and $\mathcal{D}_{val}$. For $\mathcal{D}_{opt}$, we collect optimization feedback $\mathcal{F}_{opt}(\mathcal{G}')=\{(\hat{\pi}_i,\hat{r}_i,\mathcal{O}_i)\mid (q_i,r_i,\pi_i)\in\mathcal{D}_{opt}\}$, where~$\hat{\pi}_i$~is~the produced tool orchestration trace, $\hat{r}_i$ is the execution result, and $\mathcal{O}_i$ denotes the node-level observations collected during execution. This feedback is passed to the optimizer for~graph~search.

For $\mathcal{D}_{val}$, we evaluate the candidate workflow by comparing each $\hat{\pi}_i$ with the validated orchestration log $\pi_i$ and comparing each $\hat{r}_i$ with the reference result $r_i$ to obtain the validation feedback $\textit{TC}(\mathcal{G}')_{\mathcal{D}_{val}}$ and $\textit{ES}(\mathcal{G}')_{\mathcal{D}_{val}}$, respectively. This feedback is used to guide the final workflow selection.


\subsection{Workflow Optimizer}\label{sec:workflow_optimizer}

Starting from the initial workflow $\mathcal{G}^0$, the workflow optimizer searches for a better tool-integrated agentic workflow through execution-guided graph search. We adopt Monte Carlo tree search (MCTS)~\cite{browne2012survey} because it naturally fits execution-guided workflow optimization, where each iteration expands one candidate graph, evaluates it through real execution, and uses the observed feedback to guide later search.
In the search tree, each node $m$ stores a tool-integrated agentic workflow $\mathcal{G}^m$, its visit count $C^m$ that records how many times the workflow $\mathcal{G}^m$ has been evaluated or passed through during backpropagation, accumulated reward $W^m$ that sums the scalar rewards propagated to $m$, mean reward $Q^m=\frac{W^m}{C^m}$ that estimates the historical quality of $\mathcal{G}^m$, and optimization feedback $\mathcal{F}_{opt}(\mathcal{G}^m)$ collected by the workflow executor. Each tree edge corresponds to applying one graph mutation to~the~parent~agentic~workflow. \hs{Our MCTS for workflow optimization consists of four key stages, \ie \textit{selection}, \textit{expansion}, \textit{simulation},~and \textit{backpropagation},}

\textbf{Selection.} In each iteration, \tool selects an MCTS node to expand from the current search tree. Starting from the root node corresponding to $\mathcal{G}^0$, we recursively descend the search tree until \hs{reaching an expandable node $m^\star$}. To keep the branching factor manageable, each tree node is allowed to generate at most three children in our implementation. \hs{When \tool reaches a node $m$ that has generated fewer than three children, the selection process stops, and $m$ is selected as $m^\star$ for expansion.} Otherwise, \tool chooses one of its children according to a selection score that balances historical execution quality and exploration. \hs{For each child node $m^c$ of $m$,} the selection score $\mathrm{Score}(m^c)$ is defined by Eq.~\ref{eq:selection-score},
\begin{equation}
    \label{eq:selection-score}
    \scriptsize
    \mathrm{Score}(m^c) =
    \begin{cases}
    +\infty, & C^{m^c} = 0, \\
    Q^{m^c} + \lambda \sqrt{\dfrac{\ln(C^m+1)}{C^{m^c}+1}}, & C^{m^c} > 0.
    \end{cases}
\end{equation}
where $Q^{m^c}$ is the mean reward of $m^c$, $C^m$ and $C^{m^c}$ denote the visit counts of the parent and child nodes, and $\lambda$ is the current exploration coefficient. The second term gives higher scores to less visited children, preventing the search from prematurely focusing on a small set of candidates. The coefficient $\lambda$ is adjusted by a cosine annealing schedule~\cite{loshchilov10sgdr}, encouraging~broader exploration in early iterations and gradually shifting the search toward candidates with better scalar rewards. \hs{We select the child node with the highest score as~$m^\star$ for~expansion.}

\textbf{Expansion.}
\hs{After selecting an expandable node $m^\star$}, we expand it by generating a new child workflow from $\mathcal{G}^{m^\star}$.~The expansion is guided by an LLM-based optimizer that analyzes the optimization feedback $\mathcal{F}_{opt}(\mathcal{G}^{m^\star})$ collected by the workflow executor. Recall that each feedback item $(\hat{\pi}_i,\hat{r}_i,\mathcal{O}_i)$ contains the produced orchestration trace, the execution result, and~node-level observations for each query $q_i \in \mathcal{D}_{opt}$. Thus, we~can~compare $\hat{\pi}_i$ with the validated orchestration log $\pi_i$ to identify tool-invocation mismatches, compare $\hat{r}_i$ with $r_i$ to identify execution failures, and inspect $\mathcal{O}_i$ to find problematic~workflow~nodes.

Based on these feedback signals, the LLM-based optimizer proposes a workflow refinement plan. The plan may contain one or multiple graph edits denoted as $\Delta^{m^\star}=\{o_1,o_2,\ldots,o_L\}$, where each operation $o_l$ is a graph edit applied to $\mathcal{G}^{m^\star}$.~Applying the refinement plan produces a new candidate workflow $\mathcal{G}'=\Delta^{m^\star}(\mathcal{G}^{m^\star})$. The operation set includes four types of edits. \textit{Node-level edits} add, remove, or replace LLM nodes and tool-calling nodes. \textit{Edge-level edits} add, remove, or reconnect data flow dependencies. \textit{Control-flow edits} introduce or refine conditional branches, parallel branches, and retry loops. \textit{Prompt-level edits} refine the system prompts of LLM nodes.

\hs{For example, tool-invocation mismatches may lead to inserting or refining tool-calling nodes, while invalid tool arguments may lead to revising the argument generation prompt. Wrong branch activation may lead to refining conditional branches or retry loops. When the tool trace is mostly correct but the final execution score remains low, the optimizer may refine the prompt for the LLM node for response synthesis.} These edits can be combined in one refinement plan when the feedback indicates coupled problems. To reduce the cost and instability of generating prompts for newly added LLM nodes, \tool maintains a small library of predefined LLM-node prompt templates. These templates cover common node roles in agentic workflows, including planning, extracting, reasoning, verifying, and synthesizing. When a refinement plan inserts or replaces an LLM node, \tool selects the corresponding prompt template according to the intended node role, and fills it with the current workflow context and execution feedback. \hs{All prompt templates for LLM nodes along with the prompts for the LLM-based optimizer are provided~on~our~replication~website~\cite{website}.}

\hs{Finally, \tool applies the refinement plan to $\mathcal{G}^{m^{\star}}$, and the resulting candidate workflow $\mathcal{G}'$ is added as a new child $m'$ of $m^\star$ and passed to the workflow executor for execution.}

\textbf{Simulation.}
After expansion, we invoke the workflow executor to run the new candidate workflow $\mathcal{G}'$, obtaining the execution feedback $\mathcal{F}_{opt}(\mathcal{G}')$. Since MCTS requires a scalar reward for backpropagation, we calculate the tool invocation correctness $\textit{TC}(\mathcal{G}')_{\mathcal{D}_{opt}}$ and the execution score $\textit{ES}(\mathcal{G}')_{\mathcal{D}_{opt}}$ on $\mathcal{D}_{opt}$, and combine them into a scalar~reward~by~Eq.~\ref{eq:reward}.
\begin{equation}
    \label{eq:reward}
    \small
    \mathrm{R}(\mathcal{G}')_{\mathcal{D}_{opt}}=\frac{1}{2}\left(
    \textit{TC}(\mathcal{G}')_{\mathcal{D}_{opt}}
    +
    \textit{ES}(\mathcal{G}')_{\mathcal{D}_{opt}}
    \right)
\end{equation}
The reward $\mathrm{R}(\mathcal{G}')_{\mathcal{D}_{opt}}$ is used for backpropagation, while $\mathcal{F}_{opt}(\mathcal{G}')$ is cached in the new child node for later expansion. During optimization, \tool also records the $\textit{TC}(\mathcal{G}')_{\mathcal{D}_{val}}$ and $\textit{ES}(\mathcal{G}')_{\mathcal{D}_{val}}$, and computes the validation score $\mathrm{R}(\mathcal{G}')_{\mathcal{D}_{val}}$ in the same form as Eq.~\ref{eq:reward}. The $\mathrm{R}(\mathcal{G}')_{\mathcal{D}_{val}}$ is used only for candidate selection and is not used to update MCTS statistics.

\textbf{Backpropagation.}
After simulation, \tool propagates the scalar reward $\mathrm{R}(\mathcal{G}')_{\mathcal{D}_{opt}}$ \hs{from the expanded child~node back to the root node along the selected path.} For each node $m^i$ on this path, we update $C^{m^i}$, $W^{m^i}$, and $Q^{m^i}$ by~Eq.~\ref{eq:backpropagation}.
\begin{equation}
    \label{eq:backpropagation}
    \small
    C^{m^i} \leftarrow C^{m^i}+1, \\
    W^{m^i} \leftarrow W^{m^i}+\mathrm{R}(\mathcal{G}')_{\mathcal{D}_{opt}}, \\
    Q^{m^i} \leftarrow \frac{W^{m^i}}{C^{m^i}}.
\end{equation}
These updated statistics affect the selection in later iterations.

\begin{algorithm}[t]
    \LinesNumbered
    \footnotesize
    \caption{\textsc{FlowScout} Workflow Optimizer}
    \label{alg:workflow-optimizer}
    \KwIn{Initial workflow $\mathcal{G}^{0}$, optimization set $\mathcal{D}_{opt}$,
    validation set $\mathcal{D}_{val}$, maximum iterations $J$, patience $P$}
    \KwOut{Optimized workflow $\mathcal{G}^{*}$}

    Initialize root node of the tree $\mathcal{S}$ with $\mathcal{G}^{0}$ and $\mathcal{F}_{opt}(\mathcal{G}^{0})$\label{alg:init-root}\;
    $\mathcal{G}^{*} \gets \mathcal{G}^{0}$,
    $\mathrm{R}^{*}_{val} \gets \mathrm{R}(\mathcal{G}^{0})_{\mathcal{D}_{val}}$,
    $p \gets 0$\label{alg:init-best}\;

    \For{$j \gets 1$ \KwTo $J$}{
        $m^\star \gets \mathrm{Select}(\mathcal{S})$\label{alg:select}\;
        $\mathcal{G}' \gets \mathrm{Expand}(m^\star, \mathcal{S}.m^\star.\mathcal{F}_{opt}(\mathcal{G}^{m^\star}))$\label{alg:expand}\;

        $\mathrm{R}(\mathcal{G}')_{\mathcal{D}_{opt}}$, $\mathcal{S}.m'.\mathcal{F}_{opt}(\mathcal{G}')$ $\gets \mathrm{Simulate}(\mathcal{G}'$, $\mathcal{D}_{opt})$\label{alg:simulate1}\;
        $\mathrm{R}(\mathcal{G}')_{\mathcal{D}_{val}} \gets \mathrm{Simulate}(\mathcal{G}'$, $\mathcal{D}_{val})$ \label{alg:simulate2}\;

        $\mathrm{Backpropagate}(\mathcal{S}.m', \mathrm{R}(\mathcal{G}')_{\mathcal{D}_{opt}})$\label{alg:backpropagate}\;

        \If{$\mathrm{R}(\mathcal{G}')_{\mathcal{D}_{val}} > \mathrm{R}^{*}_{val}$}{
            $\mathcal{G}^{*} \gets \mathcal{G}'$\label{alg:update-workflow}\;
            $\mathrm{R}^{*}_{val} \gets \mathrm{R}(\mathcal{G}')_{\mathcal{D}_{val}}$\label{alg:update-score}\;
            $p \gets 0$\label{alg:reset-patience}\;
        }
        \Else{
            $p \gets p+1$\label{alg:update-patience}\;
        }

        \If{$p \ge P$}{
            \textbf{break}\label{alg:early-stop}\;
        }
    }

    \Return{$\mathcal{G}^{*}$}\label{alg:return}\;
\end{algorithm}

\textbf{Optimization Procedure.}
Algorithm~\ref{alg:workflow-optimizer} summarizes the optimization procedure. \tool first initializes the search tree $\mathcal{S}$ with the initial workflow $\mathcal{G}^{0}$ and its execution feedback $\mathcal{F}_{opt}(\mathcal{G}^{0})$, and sets the initial workflow as the current best \hs{candidate workflow} (Lines~\ref{alg:init-root}-\ref{alg:init-best}). In each iteration, \tool selects one node from the current search tree, and expands it into a new child workflow using the cached execution feedback (Lines~\ref{alg:select}-\ref{alg:expand}). The new workflow is then simulated on the optimization set to obtain the scalar reward for MCTS update and the execution feedback for later expansion, and is also evaluated on the validation set to obtain its validation score (Lines~\ref{alg:simulate1}-\ref{alg:simulate2}). The scalar reward is propagated along the selected path to update the MCTS statistics (Line~\ref{alg:backpropagate}). If the validation score of the new workflow is higher than the best score observed so far, \tool updates the best candidate and resets the early-stopping counter $p$ (Lines~\ref{alg:update-workflow}-\ref{alg:reset-patience}); otherwise, the counter is increased (Line~\ref{alg:update-patience}). The search terminates when the counter reaches the patience threshold $P$ (Line~\ref{alg:early-stop}) or the maximum number of iterations $J$ is reached, and finally returns the best validation candidate as $\mathcal{G}^{*}$ (Line~\ref{alg:return}). Following~\cite{zhang2025aflow}, we set the maximum number of iterations to 30. To reduce unnecessary search cost, we further apply early stopping when the validation score does not improve for 5 iterations.

\section{Evaluation}\label{sec:evaluation}
We implement a prototype of \tool with \hs{7,768} lines of Python code. To evaluate the effectiveness and efficiency of \tool, we design the following four research questions.

\begin{itemize}[leftmargin=*]
    \item \textbf{RQ1 Effectiveness Evaluation.} How effective is \tool in generating tool-integrated agentic workflows?
    \item \textbf{RQ2 Efficiency Evaluation.} How efficient is \tool in generating tool-integrated agentic workflows?
    \item \textbf{RQ3 Ablation Study.} \hs{How do the workflow miner and MCTS contribute to the final effectiveness and efficiency?}
    \item \textbf{RQ4 Generalization Evaluation.} \hs{Can the tool-integrated agentic workflows generated by \tool generalize when deployed at runtime with tools and LLM backbones different from those used during workflow generation?}
\end{itemize}

\subsection{Evaluation Setup}
\label{sec:evaluation_setup}

\textbf{Dataset.} We select four task domains, \ie finance, sports, travel, and weather from ToolBench~\cite{qin2024toolllm}, which is a widely used benchmark including user queries, available tools (\ie remote APIs), reference execution results, and validated tool orchestration logs for task completion. We first filter out the tools yielding empty responses due to unavailable services. Then, for each task domain, we divide the remaining tools into two disjoint sets, denoted as \textit{Seen-Tools} and \textit{Unseen-Tools}, where each seen tool is matched with an unseen tool that provides similar functionality. The seen tools are used for agentic workflow generation, while the unseen tools are reserved for evaluating whether the generated workflows can generalize to functionally similar but previously unseen tools. Based on the validated tool orchestration logs, we associate each tool with the user queries that invoke it during task completion. For the seen tools, we collect the associated user queries along with their reference execution results and validated tool orchestration logs in each task domain, and split them into optimization, validation, and test sets with a ratio of 5:2:3, denoted as $\mathcal{Q}_{opt}$, $\mathcal{Q}_{val}$, and $\mathcal{Q}_{test}$. The optimization set is used to guide workflow generation, the validation set is used to evaluate candidate workflows during the search, and the test set is held out for the final effectiveness evaluation. For the unseen tools, we collect the same number of associated user queries as $\mathcal{Q}_{gene}$ along with reference execution results separately, and use them only for the generalization evaluation, so that the seen and unseen test sets are of equal size for a fair comparison. The final dataset statistics are shown~in~Table~\ref{tab:dataset}.

\begin{table}[t]
    \centering
    \caption{Statistics of the Evaluation Dataset}
    \label{tab:dataset}
    \begin{adjustbox}{width=\linewidth}
    \begin{tabular}{lcccccc}
        \toprule
        \multirow{2}{*}{Domain} 
        & \multicolumn{4}{c}{\textit{Seen-Tools}} 
        & \multicolumn{2}{c}{\textit{Unseen-Tools}} \\
        \cmidrule(lr){2-5} \cmidrule(lr){6-7}
        & \#Tools
        & \#$\mathcal{Q}_{opt}$ 
        & \#$\mathcal{Q}_{val}$ 
        & \#$\mathcal{Q}_{test}$ 
        & \#Tools 
        & \#$\mathcal{Q}_{gene}$ \\
        \midrule
        Finance & 341 & 467 & 187 & 280 & 341 & 280 \\
        Sports  & 288 & 253 & 101 & 152 & 288 & 152 \\
        Travel  & 101 & 217 & 87 & 130 & 101 & 130 \\
        Weather & 48 & 83 & 33 & 50 & 48 & 50 \\
        \bottomrule
    \end{tabular}
    \end{adjustbox}
\end{table}

\textbf{Baselines.} We compare \tool with three representative baselines, \ie PM4Py~\cite{van2004workflow}, ReAct~\cite{yao2023react}, and AFlow~\cite{zhang2025aflow}. PM4Py is a representative workflow mining framework that discovers process models from event logs. We include PM4Py to examine whether traditional workflow mining techniques can derive effective workflows from validated tool orchestration logs. Specifically, we adapt each tool invocation in the logs as an activity and use the discovered process model as the workflow structure for task execution. ReAct is a widely used LLM agent paradigm that interleaves reasoning and acting. It serves as a standard agent baseline that solves tool-use tasks without explicitly constructing reusable workflow structures. AFlow is a recent automatic agentic workflow generation approach that searches for workflows from historical task-solving records. We include AFlow as the most related baseline, since it also automates workflow generation but mainly constructs LLM-centric workflows. For fair comparison, all approaches are evaluated on the same task domains, available tools, and test queries. Approaches requiring historical records are provided with the same seen-tool data, including user queries, reference execution results, and validated tool orchestration logs. \hs{Both AFlow and \tool use GPT-4o~\cite{gpt4o} as the optimization LLM for workflow construction. The workflows generated by AFlow and \tool, together with the ReAct agent, use GPT-4o-mini~\cite{gpt4omini} as the execution LLM backbone and have access to the same tool documentation.} For all baselines, we only align the input and output interfaces of their official replication artifacts to fit our evaluation dataset.

\textbf{RQ Setup.} 
For \textbf{RQ1}, we use the optimization set $\mathcal{Q}_{opt}$ and validation set $\mathcal{Q}_{val}$ to generate workflows with PM4Py, AFlow, and \tool. We then evaluate the generated workflows, together with the ReAct agent, on the test set $\mathcal{Q}_{test}$. We report the average $\textit{TC}$ and $\textit{ES}$ across different task domains, and compare \tool with the baselines. We further run the workflows generated by AFlow and \tool for 10 times, as well as the ReAct agent, on 50 queries selected randomly in $\mathcal{Q}_{test}$ for each task domain, and compare the coefficient of variation of $\textit{ES}$ to evaluate their execution stability.

\hs{For \textbf{RQ2}, we evaluate the efficiency of \tool from both offline workflow generation and runtime execution. For offline generation, we plot the validation score of candidate workflows in 30 iterations to show how the quality of workflows evolves during the search process. For runtime execution, we measure the average execution time of the workflow generated by \tool on $\mathcal{Q}_{test}$, and compare it with that of the ReAct agent and the workflow generated~by~AFlow.}

\hs{For \textbf{RQ3}, we conduct ablation studies on the finance domain to examine the contribution of the workflow miner and the MCTS. We construct three variants of \tool, denoted as \tool-\textsc{NoMCTS}, \tool-\textsc{NoMiner}, and \tool-\textsc{Beam}, respectively. \tool-\textsc{NoMCTS} removes the MCTS-based workflow optimizer, and directly uses the workflow mined from the historical task-solving records without further optimization. \tool-\textsc{NoMiner} removes the workflow miner from \tool, and searches for workflows from an empty skeleton using MCTS-based workflow optimizer. \tool-\textsc{Beam} replaces MCTS with beam search~\cite{russell2021artificial}, which keeps the top-3 workflow candidates at each search step. For fair comparison, \tool-\textsc{NoMiner}, \tool-\textsc{Beam} and \tool are allowed to generate 30 candidate workflows in total. We compare the \textit{TC} and \textit{ES} of their best candidate workflows to evaluate effectiveness, and report the total generation time cost to evaluate efficiency.}

\hs{For \textbf{RQ4}, we evaluate the workflow generated by \tool from two aspects, \ie tool-set replacement and LLM-backbone replacement, on the finance domain compared with the ReAct agent and the workflow generated by AFlow. For tool-set generalization, the tool-integrated agentic workflow is generated using \textit{Seen-Tools}, and then evaluated after replacing the available tools with \textit{Unseen-Tools}. For LLM backbone generalization, we keep the generated workflow topology unchanged, and replace the execution LLM backbone with DeepSeek-V3.2~\cite{deepseek2025deepseekv32} and Qwen3-32B~\cite{yang2025qwen3}, respectively. We report $\textit{TC}$ and $\textit{ES}$ to measure whether each approach can still support effective tool invocation and task completion.}

\textbf{Environment.} We conduct all the experiments on Ubuntu 20.04.4 LTS servers with 4 NVIDIA GeForce RTX 3090 GPUs, Intel(R) Xeon(R) Silver 4310 @ 2.10GHz and 128GB memory.

\subsection{Effectiveness Evaluation (RQ1)}\label{sec:evaluation_rq1}

\begin{table}[t]
\centering
\caption{Results of the General Effectiveness}
\label{tab:rq1_effectiveness}
\setlength{\tabcolsep}{4pt}
\resizebox{\columnwidth}{!}{%
\begin{tabular}{lcccccc}
\toprule
\multirow{2}{*}{Approach} & \multirow{2}{*}{Metric} & \multicolumn{5}{c}{Task Domain} \\
\cmidrule(lr){3-7}
& & Finance & Sports & Travel & Weather & \cellcolor{gray!15} Avg. \\
\midrule
\multirow{2}{*}{PM4Py}
& \textit{TC} & 0.0550 & 0.0353 & 0.0358 & 0.2926 &\cellcolor{gray!15} 0.1047 \\
& \textit{ES} & 0.1530 & 0.1715 & 0.0620 & 0.2680 &\cellcolor{gray!15} 0.1636 \\
\midrule
\multirow{2}{*}{ReAct}
& \textit{TC} & 0.5115 & 0.3501 & 0.5046 & 0.1939 &\cellcolor{gray!15} 0.3900 \\
& \textit{ES} & 0.6113 & 0.6418 & 0.6184 & 0.6376 &\cellcolor{gray!15} 0.6273 \\
\midrule
\multirow{2}{*}{AFlow}
& \textit{TC} & -- & -- & -- & -- &\cellcolor{gray!15} -- \\
& \textit{ES} & 0.6182 & 0.6007 & 0.6848 & 0.5651 &\cellcolor{gray!15} 0.6172 \\
\midrule
\multirow{2}{*}{\tool}
& \textit{TC} & \textbf{0.6583} & \textbf{0.7160} & \textbf{0.8028} & \textbf{0.8289} &\cellcolor{gray!15} \textbf{0.7515} \\
& \textit{ES} & \textbf{0.7349} & \textbf{0.6944} & \textbf{0.7528} & \textbf{0.7701} &\cellcolor{gray!15} \textbf{0.7381} \\
\bottomrule
\end{tabular}%
}
\vspace{0.3em}
\footnotesize
\end{table}

\textbf{Overall Effectiveness.}
Table~\ref{tab:rq1_effectiveness} reports the effectiveness results of \tool and the baselines on the test set. We do not report \textit{TC} for AFlow because its generated workflow only contains LLM nodes and does not produce real-tool invocations. Overall, \tool consistently achieves the best performance across all four domains in terms of both tool-invocation correctness (\textit{TC}) and execution score (\textit{ES}). 

PM4Py performs markedly worse than all LLM-based approaches. Although process mining can extract tool orchestration patterns from historical task-solving records, directly translating these patterns into executable workflows provides limited support for query understanding, tool selection, argument construction, and result synthesis. For fair comparison, we mainly compare \tool with LLM-based approaches.

With respect to \textit{TC}, \tool improves the average \textit{TC} over ReAct by \hs{92.69\%}. The improvement is observed in every domain, ranging from \hs{28.70\%} in the finance domain to \hs{327.49\%} in the weather domain. In particular, \tool more than doubles the \textit{TC} of ReAct in the sports domain and achieves over four times its \textit{TC} in the weather domain, demonstrating that the generated tool-integrated agentic workflows substantially improve the accuracy of tool selection and orchestration.

With respect to \textit{ES}, \tool improves the average \textit{ES} by \hs{17.66\%} over ReAct and \hs{19.59\%} over AFlow. \tool also consistently surpasses the best-performing baseline in each domain, with relative improvements of \hs{18.88\%}, \hs{8.20\%}, \hs{9.93\%}, and \hs{20.78\%} in the finance, sports, travel, and weather domains, respectively. \tool outperforms both baselines in all cases, indicating that it can generate workflows that are both more effective and less sensitive to domain-specific tasks.

Overall, these results demonstrate that \tool generates workflows with more accurate tool invocation and better end-to-end task-solving performance than both general-purpose LLM agents and existing workflow-generation approaches.

\begin{figure}[t]
    \centering
    \includegraphics[width=\linewidth]{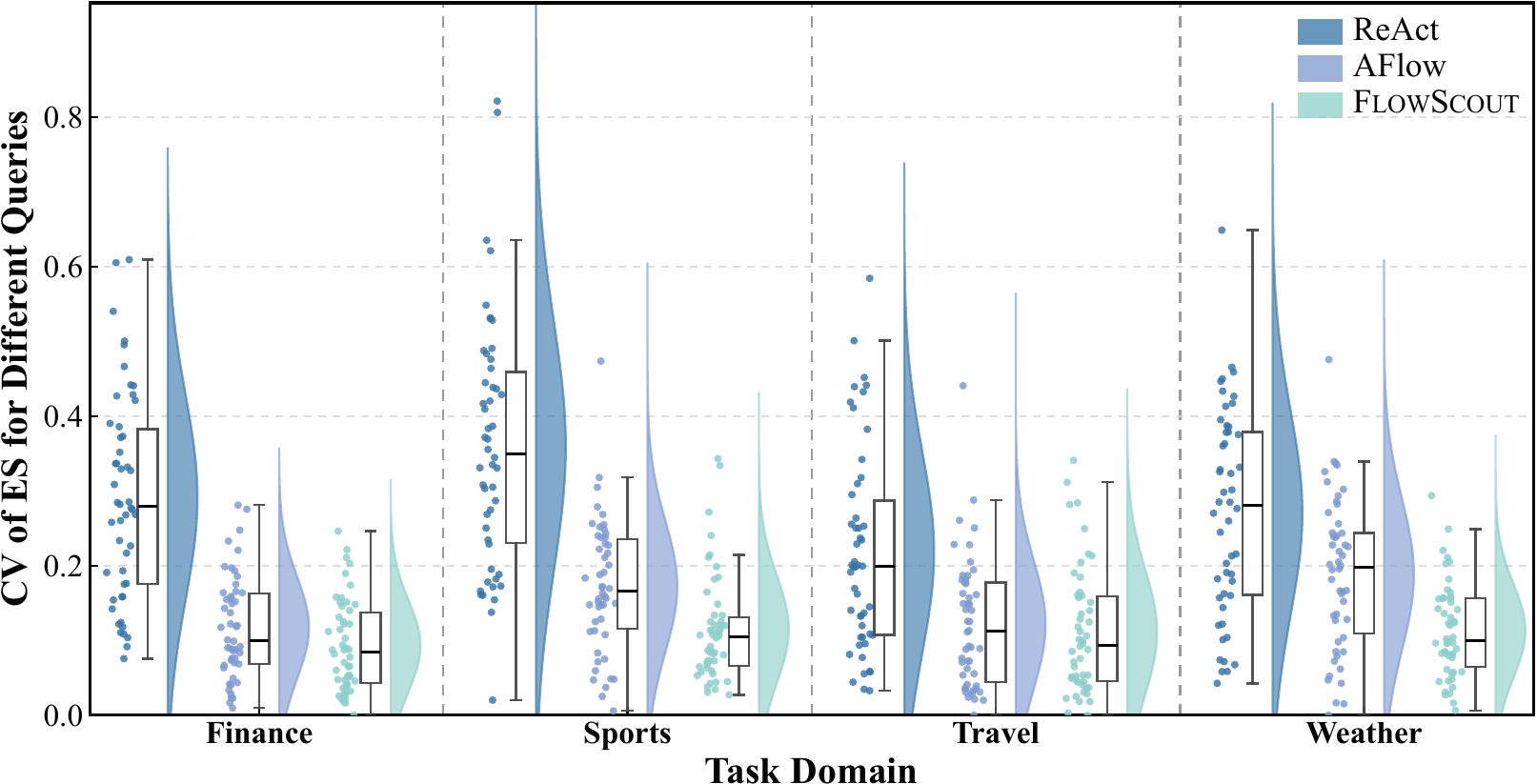}
    \caption{Results of the Execution Stability}
    \label{fig:stability}
\end{figure}

\textbf{Stability Analysis.}
We further evaluate execution stability by repeatedly running each approach 10 times on the same~test queries and computing the coefficient of variation (CV) of \textit{ES} for each query. A lower CV indicates that the approach has more consistent execution quality across repeated~runs. As shown in Fig.~\ref{fig:stability}, \tool achieves the lowest~average CV across domains. Overall, it reduces the~average~CV~by \hs{62.09\%} and \hs{27.90\%} compared with ReAct and AFlow, respectively.

Compared with ReAct, the improvement is particularly substantial in the finance and sports domains, where \tool reduces the average CV by \hs{68.14\%} and \hs{68.03\%}, respectively. These domains involve relatively large tool spaces and more possible tool combinations, making unconstrained planning more susceptible to different tool selections and invocation orders across runs. This consistent reduction in CV across all domains indicates that explicitly structuring workflow benefits in complex scenarios. Compared with AFlow, \tool reduces the average CV by \hs{20.75\%}, \hs{33.22\%}, \hs{8.31\%}, and \hs{40.04\%} in the finance, sports, travel, and weather domains, respectively. The box-plot distributions show that \tool generally exhibits narrower interquartile ranges and fewer high-variation cases. Although AFlow generates reusable workflow structures, its workflows are primarily composed of LLM-centric operators without actual tool invocations. In contrast, \tool integrates tool-calling nodes and their dependencies explicitly, and refines them using execution feedback. This reduces the number of runtime decisions left to unstable LLM planning which may lead to hallucinations and consequently improves the reproducibility of execution~outcomes.

\textbf{Breakdown Analysis.}
The observed ineffectiveness and instability can be attributed to uncertainty introduced at three stages of execution, \ie LLM planning, tool invocation, and response synthesis. For ReAct, all three stages are dynamically determined at runtime. Variations in LLM generation may lead the agent to decompose the same query differently, select different tools, alter the invocation order, or terminate at different points, producing substantially different final execution scores. AFlow reduces part of this variation by providing reusable agentic workflows. However, because tool invocation is not represented and the agentic workflows generated by AFlow rely heavily on LLM nodes, AFlow is generally more stable than ReAct but still exhibits noticeable variation.

\tool reduces both ineffectiveness and instability by constructing the workflow topology and explicitly modeling tool invocations. However, the generated tool-integrated agentic workflows are not fully deterministic or universally optimal. Their effectiveness remains limited by insufficient coverage of real world queries, and errors in LLM-based query interpretation, argument generation, and response synthesis. Meanwhile, stochastic decisions within LLM nodes, varying intermediate outputs, and occasional tool failures can still lead to inconsistent execution results across repeated runs.

\begin{figure*}[t]
    \centering
    \includegraphics[width=0.9\linewidth]{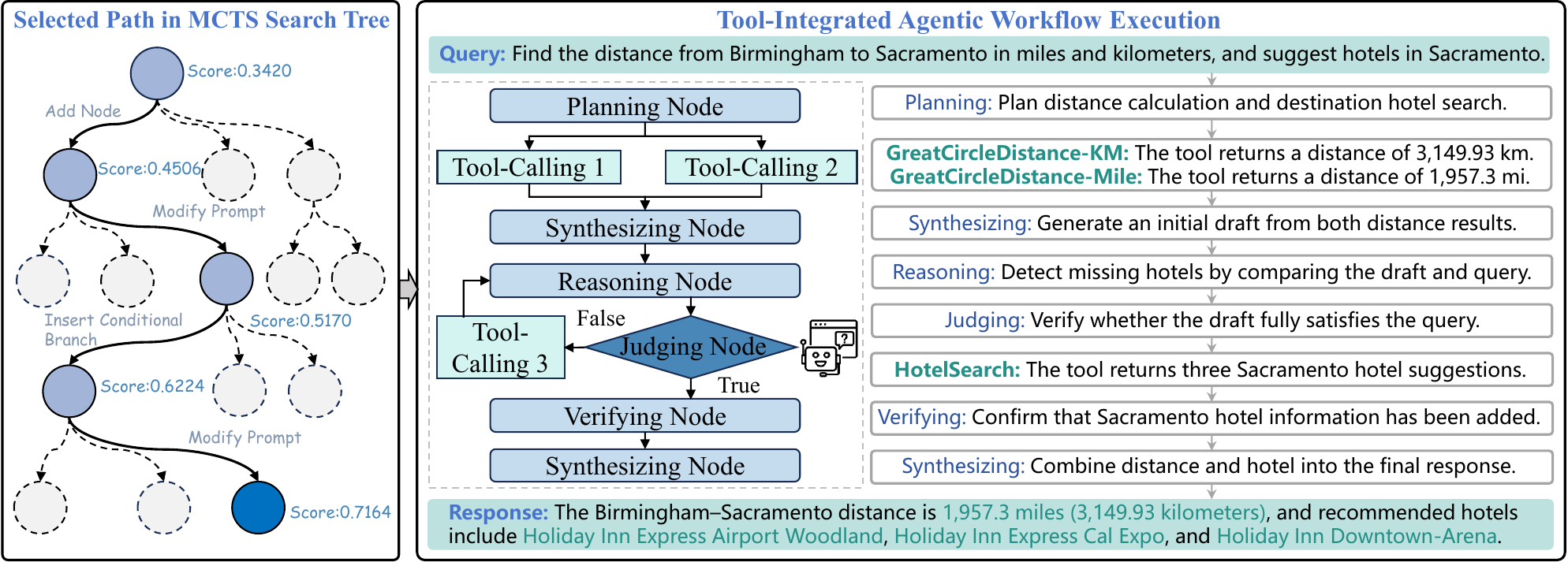}
    \caption{Examples of the Workflow Search Process and the Tool-Integrated Agentic Workflow in the Travel Domain}
    \label{fig:case_study}
\end{figure*}

\textbf{Case Study.} Fig.~\ref{fig:case_study} illustrates an abstract workflow search process in the travel domain and the execution of the resulting tool-integrated workflow. The left part shows the selected path in the MCTS search tree, where each node represents a candidate workflow and each edge represents a workflow mutation. Through operations such as adding nodes, inserting conditional branches, and modifying prompts, \tool progressively improves the validation score from \hs{0.3420}~to~\hs{0.7164}.

The right part shows the execution of the tool-integrated agentic workflow on a query that requests both the distance from Birmingham to Sacramento and hotel recommendations in Sacramento. The workflow first decomposes the query into distance calculation and hotel recommendation subtasks. It then invokes \texttt{[GreatCircleDistance-KM]} and \texttt{[GreatCircleDistance-Mile]} to obtain the distance in kilometers and miles, respectively. When the intermediate context is found to omit the requested hotel information by the reasoning node, the judging node activates the optional branch and invokes \texttt{[HotelSearch]} to retrieve hotels in Sacramento. Finally, the synthesizing node integrates the outputs of the three tools into a final response.

    \textit{\textbf{Summary.}} 
    \tool improves the tool invocation correctness by at least \hs{92.69\%} and execution score by at least \hs{17.66\%} with better stability, compared with baselines. 

\subsection{Efficiency Evaluation (RQ2)}\label{sec:evaluation_rq2}

\begin{figure}[t]
    \centering
    \includegraphics[width=\linewidth]{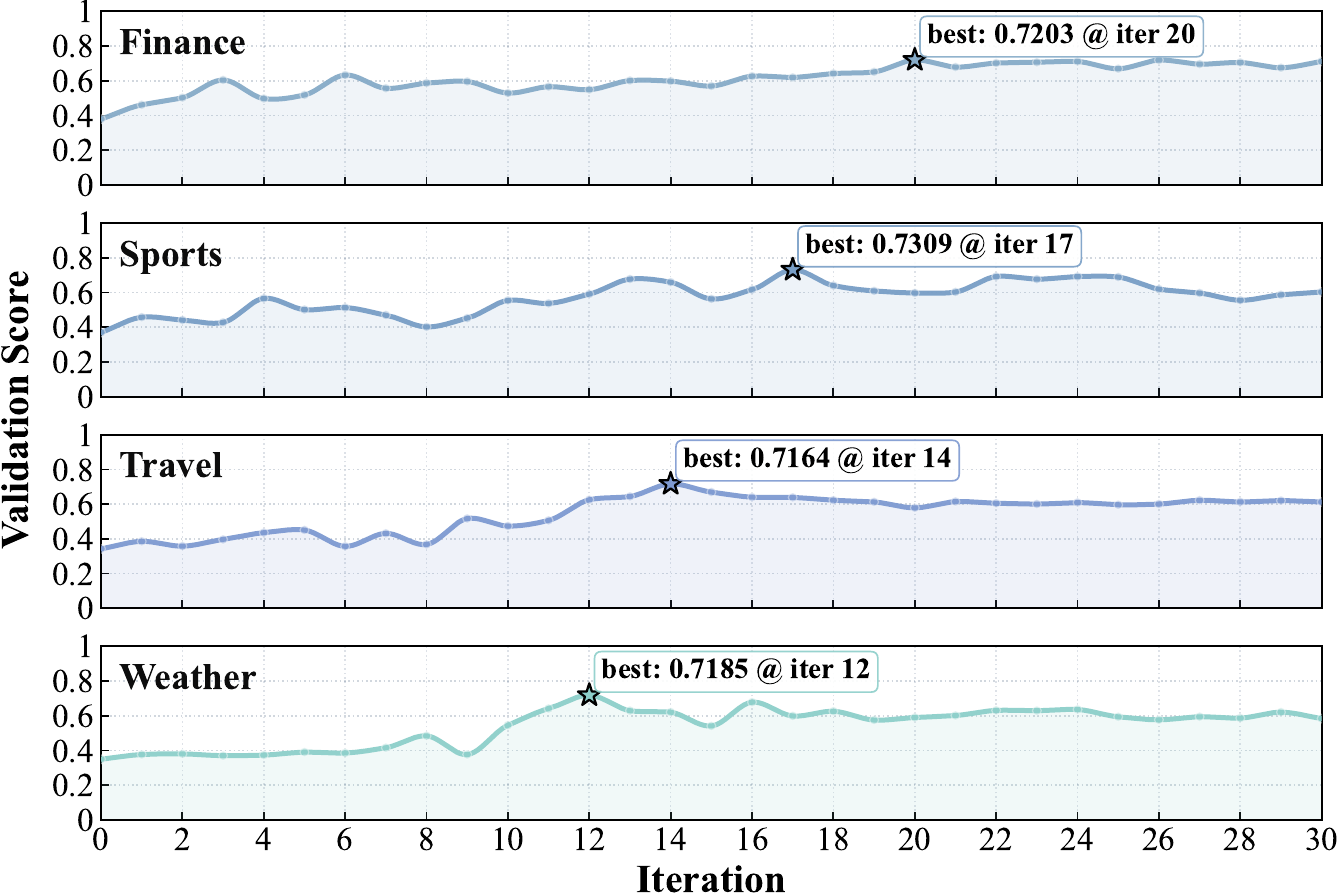}
    \caption{Validation Scores during the Search Process}
    \label{fig:optimization_curve}
\end{figure}

\textbf{Offline Workflow Generation.}
Fig.~\ref{fig:optimization_curve} shows the validation scores of candidate workflows over 30 optimization iterations. Across all four domains, \tool finds its best workflow within 20 iterations, indicating that execution-guided search can obtain effective workflows under a bounded optimization budget. Compared with the initial workflows, the best candidates improve the validation scores by \hs{92.20\%}, \hs{98.13\%}, \hs{109.47\%}, and \hs{105.58\%} in the finance, sports, travel, and weather domains, respectively, demonstrating the benefit of iterative workflow optimization. Besides, these best candidates appear at iterations 20, 17, 14, and 12, respectively, where the ordering is consistent with the tool-space sizes of these four task domains. This result suggests that larger tool spaces may require more iterations to explore the optimal tool-integrated~agentic workflow.

\textbf{Runtime Execution.}
At runtime, the agentic workflow generated by \tool requires 22.80 seconds per query, on average, compared with 11.44 seconds for the ReAct agent and 18.37 seconds for the agentic workflow generated by AFlow. The overhead is more pronounced relative to the ReAct agent, which performs planning and tool invocation within a comparatively compact interaction loop, while \tool may involve multiple specialized LLM nodes for planning, reasoning, judging, and synthesizing intermediate results. These nodes reduce unconstrained runtime decisions, but also introduce additional runtime cost. In contrast, the smaller gap between \tool and AFlow indicates that part of the runtime cost is common to workflow-based execution, while \tool introduces time overhead for tool invocation compared with AFlow. Considering that \tool improves both general effectiveness and execution stability over the baselines, the observed runtime overhead represents a trade-off between execution efficiency and workflow reliability.

    \textit{\textbf{Summary.}} \tool finds effective workflows within a limited offline search budget, with at least \hs{24.12\%} higher runtime cost than baselines for improved effectiveness.

\subsection{Ablation Study (RQ3)}\label{sec:evaluation_rq3}

\begin{table}[t]
  \centering
  \caption{Results of the Ablation Study}
  \label{tab:rq3_ablation}
  \begin{adjustbox}{width=\linewidth}
  \begin{tabular}{lcccccc}
  \toprule
  & Miner       & MCTS       & Beam       & $\textit{TC}$ & $\textit{ES}$ & Time (h) \\
  \midrule
  \tool-\textsc{NoMCTS} & \checkmark & & & 0.4598 & 0.2118 & 0.01 \\
  \tool-\textsc{NoMiner}& & \checkmark & & 0.4985 & 0.5621 & 10.9 \\
  \tool-\textsc{Beam} & \checkmark & & \checkmark & \textbf{0.6768} & 0.5327 & 38.88 \\
  \rowcolor{gray!15}
  \tool & \checkmark & \checkmark &            & 0.6583 & \textbf{0.7349} & 15.39 \\
  \bottomrule
  \end{tabular}
  \end{adjustbox}
\end{table}

Table~\ref{tab:rq3_ablation} reports the ablation results on the finance domain. Among all variants, \tool-\textsc{NoMCTS} shows the largest performance degradation compared with \tool, reducing \textit{TC} and \textit{ES} by \hs{30.15\%} and \hs{71.18\%}, respectively. This indicates that the mined workflow provides a useful tool coordination skeleton, but still requires further optimization to become an effective tool-integrated agentic workflow.

Removing the workflow miner also reduces performance. Compared with \tool, \tool-\textsc{NoMiner} reduces \textit{TC} by \hs{24.27\%} and \textit{ES} by \hs{23.51\%}. Without the mined skeleton, MCTS cannot discover an optimal candidate workflow within the same candidate budget. The miner therefore provides a warm start that improves both search quality and efficiency.

Replacing MCTS with beam search yields a \hs{2.81\%} higher \textit{TC}, but its \textit{ES} is \hs{27.51\%} lower than that of \tool. Beam search retains only the top-performing candidates at each step and may prematurely discard workflows that can lead to better optimization results. In contrast, MCTS models each candidate agentic workflow as a node in the search tree and explores the entire search tree for better optimization results.

Regarding the time cost, \tool-\textsc{NoMCTS} is the fastest because it performs no iterative optimization, but its effectiveness is substantially lower. Under the same candidate budget, \tool takes \hs{15.39} hours, which is \hs{41.19\%} more than \tool-\textsc{NoMiner} and \hs{60.42\%} less than \tool-\textsc{Beam}. The high cost of beam search arises from maintaining and expanding multiple candidates at each search step. These results show that both workflow mining and MCTS contribute to the final effectiveness and efficiency.

  \textit{\textbf{Summary.}} The workflow miner provides an effective initialization, while MCTS further improves TC and ES over the mined-only workflow by 43.17\% and 246.98\%. Compared with beam search, MCTS achieves substantially higher agentic workflow quality with approximately half the time for search.

\subsection{Generalization Evaluation (RQ4)}\label{sec:evaluation_rq4}

\begin{table}[t]
\centering
\caption{Results of Tool Generalization}
\label{tab:rq4_tool_generalization}
\begin{tabular}{lcccc}
\toprule
 & Metric & ReAct & AFlow & \tool \\
\midrule
\multirow{2}{*}{Seen-Tools}   & \textit{TC} & 0.5115 & -- & \textbf{0.6583} \\
                             & \textit{ES} & 0.6113 & 0.6182 & \textbf{0.7349} \\
\midrule
\multirow{2}{*}{Unseen-Tools} & \textit{TC} & 0.4574 & -- & \textbf{0.5660} \\
                             & \textit{ES} & 0.6050 & 0.5874 & \textbf{0.6751} \\
\bottomrule
\end{tabular}
\end{table}

\textbf{Tool-Set Generalization.}
Table~\ref{tab:rq4_tool_generalization} reports the results after replacing the tools used during workflow generation with \textit{Unseen-Tools}. Although all approaches show some degradation, \tool remains the best-performing approach on both metrics. Compared with ReAct, \tool improves \textit{TC} by \hs{23.74\%} and \textit{ES} by \hs{11.59\%} under the unseen tool set. It also outperforms AFlow by \hs{14.93\%} in \textit{ES}.

After tool-set replacement, \tool retains \hs{85.98\%} of its original \textit{TC} and \hs{91.86\%} of its original \textit{ES}. In particular, the relatively small decrease in \textit{ES} shows that the generated agentic workflow can preserve most of its task-solving capability when deployed with different available tools. Together with its continued advantage over the baselines, this result demonstrates that the generated agentic workflow is not restricted to the tools used during workflow generation. This generalization mainly stems from the separation between workflow-level orchestration and runtime tool binding. \tool preserves reusable task decomposition, control-flow, and dependency structures, while its tool-calling nodes dynamically select tools and construct arguments according to the tools available at runtime, rather than binding the workflow to specific tools.

\begin{table}[t]
\centering
\caption{Results of LLM Generalization}
\vspace{-3pt}
\label{tab:rq4_llm_generalization}
\begin{tabular}{lcccc}
\toprule
 & Metric & ReAct & AFlow & \tool \\
\midrule
\multirow{2}{*}{GPT-4o-mini}   & \textit{TC} & 0.5115 & --     & \textbf{0.6583} \\
                               & \textit{ES} & 0.6113 & 0.6182 & \textbf{0.7349} \\                      
\midrule
\multirow{2}{*}{DeepSeek-V3.2} & \textit{TC} & 0.5463 & -- & \textbf{0.5548} \\
                               & \textit{ES} & 0.5676 & 0.5158 & \textbf{0.7221} \\
\midrule
\multirow{2}{*}{Qwen3-32B}     & \textit{TC} & 0.3908 & -- & \textbf{0.5885} \\
                               & \textit{ES} & 0.3902 & 0.4638 & \textbf{0.7239} \\
\bottomrule
\end{tabular}
\vspace{-2pt}
\end{table}

\textbf{LLM-Backbone Generalization.}
Table~\ref{tab:rq4_llm_generalization} reports the results after replacing GPT-4o-mini with DeepSeek-V3.2 and Qwen3-32B while keeping the generated workflow topology unchanged. \tool consistently achieves the highest \textit{TC} and \textit{ES} across both replacement LLMs. With DeepSeek-V3.2, \tool outperforms ReAct by \hs{1.56\%} in \textit{TC} and \hs{27.22\%} in \textit{ES}, and exceeds AFlow by \hs{40.00\%} in \textit{ES}. With Qwen3-32B, the improvements over ReAct reach \hs{50.59\%} and \hs{85.52\%}, respectively, while its \textit{ES} is \hs{56.08\%} higher than that of AFlow.

After replacing the execution LLM, \tool retains \hs{84.28\%} and \hs{89.39\%} of its original \textit{TC} with DeepSeek-V3.2 and Qwen3-32B, respectively. More importantly, it retains \hs{98.26\%} and \hs{98.50\%} of its original \textit{ES}, showing that the generated workflow preserves nearly all of its capability across different LLM backbones. This is because our generated workflow explicitly specifies the roles of different LLM nodes, tool-calling positions, and their dependencies, while the execution LLM is only responsible for completing the local function of each node. Therefore, replacing the LLM backbone may affect individual decisions to some extent, but the overall task-solving capability encoded by the workflow remains reusable.

    \textit{\textbf{Summary.}} The workflows generated by \tool retain \hs{91.37\%} of their original \textit{TC} and \textit{ES} on average when deployed with different tools and LLM backbones.



\section{Threats to Validity}\label{sec:threats}

First, the selection of evaluation metrics poses a threat to validity. \textit{TC} may penalize alternative but valid tool invocations, while \textit{ES} may be affected by the bias of the LLM-based evaluator. To mitigate this threat, we evaluate all approaches using the same reference execution results, employing a widely used LLM-based evaluator with the same prompts, and jointly report both tool-level and result-level metrics.

Second, the selection and configuration of baselines pose a threat to validity, since the compared approaches adopt different workflow abstractions. To reduce this threat, we use their official replication artifacts, modify only the input and output interfaces, and align the available tools, tool documentation, execution LLM, and optimization budget whenever applicable.

Third, domain coverage, workflow complexity, and historical-record quality pose threats to validity. The four evaluated domains differ in tool-set size and query distribution, but mainly contain tool-oriented tasks with relatively short tool invocation chains with at most 6 tool invocations. Thus, the findings may not fully generalize to workflows with longer tool invocation chains. Noisy or incomplete records may affect both the quality of workflow mining and optimization. We mitigate this threat by conducting experiments on manually validated records.
 
Finally, randomness in LLM generation, workflow search, and LLM-based evaluation poses a threat to result reliability. To mitigate this threat, we independently generate and evaluate workflows in four task domains with different tool sets and query distributions, where \tool consistently outperforms the baselines. Query-level paired tests further show that these improvements are statistically significant across domains after Holm correction~\cite{holm1979simple} ($\text{p}_{\mathrm{adj}}<0.05$), with 95\% bootstrap confidence intervals excluding zero and Cohen's d $> 0.63$~\cite{cohen1988statistical}. We also repeat execution 10 times on 50 randomly selected queries per domain, and report the coefficient of variation of \textit{ES}. These cross-domain, statistical, and repeated-execution results reduce the likelihood that the observed improvements are caused by a particular domain or an individual execution~run.  
\section{Related Work}\label{sec:related-work}

Recent studies and platforms~\cite{langchain, langgraph, wu2023autogen, dify, coze} have facilitated the construction of agentic workflows. Developer-oriented frameworks such as LangChain~\cite{langchain}, LangGraph~\cite{langgraph}, and AutoGen~\cite{wu2023autogen} provide programming abstractions for~composing LLMs, tools, memory modules and control logic. These frameworks make it easier for developers to implement agentic workflows, but they still require manual programming and careful design of workflow structures. Low-code platforms such as Dify~\cite{dify} and Coze~\cite{coze} offer visual interfaces for connecting workflow components. Although these platforms reduce the engineering effort of building agentic workflows, they still rely on users to manually specify workflow structures, tool connections, and control logic based on domain knowledge.

To further reduce the manual effort, some studies have~started to explore automatic generation of LLM agents~\cite{hu2025automated, shang2025agentsquare} or agentic workflows~\cite{li2024autoflow, zhang2025aflow, zhao2026a2flow}. Hu et al.~\cite{hu2025automated} and~Shang~et~al.~\cite{shang2025agentsquare} search over modular agent designs by composing LLMs, memory modules, external tools and prompts. AutoFlow~\cite{li2024autoflow}~parses user queries and uses LLMs to decompose them into~executable workflows at the query level, relying on the task~decomposition ability of LLMs for each individual query. The most related works to ours are AFlow~\cite{zhang2025aflow} and A$^2$Flow~\cite{zhao2026a2flow}, which also use historical task-solving records, including user~queries and execution results, to~search for agentic workflows. However, their generated agentic workflows only consist of LLM nodes, where LLMs act as general-purpose modules for solving different subtasks. In contrast,~\tool models real tool invocations as workflow nodes, and constructs the topology between LLM reasoning and external tools. Guided by execution feedback on historical task-solving records, it uses Monte~Carlo tree search to generate agentic workflows that are more usable and stable. \todo{We do not compare \tool with A$^2$Flow, as it follows the general idea of AFlow and is not~publicly available.}

\section{Conclusion}\label{sec:conclusion}
We have proposed and implemented \tool, a framework for automatic domain-specific agentic workflow generation. \tool formulates the agentic workflow as a structured graph composed of LLM nodes, tool-calling nodes and dependency edges that encode control and data dependencies, and searches for the optimal workflow topology guided by historical task-solving records. Large-scale experiments have been conducted to demonstrate its effectiveness and efficiency.

\section{Data Availability}
The source code and data of our work are available at~\cite{website}.


{\footnotesize
\bibliographystyle{IEEEtranS}
\bibliography{IEEEabrv,src/reference}
}

\end{document}